%% file: acl2020.tex
\documentclass[11pt,a4paper]{article}
\usepackage[hyperref]{acl2020}
\usepackage{times}
\usepackage{latexsym}

\usepackage{microtype}

\usepackage{kotex}
\usepackage{amsmath} 
\usepackage{amssymb} 
\usepackage[toc,page]{appendix}
\usepackage{float}
\usepackage{amsthm} 
\usepackage{graphicx} 
\usepackage{array} 
\usepackage{multirow}
\usepackage{csquotes} 
\usepackage{algorithm}
\usepackage{algorithmic}
\usepackage{hyperref}
\usepackage{multicol}

\usepackage[para]{threeparttable}
\usepackage{booktabs}
\usepackage{tabularx}
\usepackage{makecell}

\graphicspath{{figs/}}

\newcommand{\dimension}[1]{\in \mathbb{R}^{#1}}

\newcommand{\Ours}{Selectively Overwriting Memory for Dialogue State Tracking}
\newcommand{\ours}{SOM-DST}

\newcommand{\SOP}{State operation predictor}
\newcommand{\sop}{state operation predictor}
\newcommand{\SVG}{Slot value generator}
\newcommand{\svg}{slot value generator}

\newcommand{\sotaa}{51.72\%}
\newcommand{\sotab}{53.01\%}

\newcommand{\ie}{i.e.}
\newcommand{\eg}{e.g.}

\newcommand{\operation}[1]{\textsc{\textbf{#1}}}

\aclfinalcopy 

\title{Efficient Dialogue State Tracking by Selectively Overwriting Memory}

\author{
    Sungdong Kim \quad Sohee Yang \quad Gyuwan Kim \quad Sang-Woo Lee \\
    Clova AI, NAVER Corp. \\
    \texttt{\{sungdong.kim, sh.yang, gyuwan.kim, sang.woo.lee\}@navercorp.com}
}

\date{}

\begin{document}
\maketitle

\input{00_abstract}

\input{01_introduction}
\input{02_related_work}
\input{03_proposed_method_clean}
\input{04_experimental_setup}
\input{05_experimental_results}

\input{06_analysis}
\input{07_conclusion}
\input{08_acknowledgements}

\bibliography{acl2020}
\bibliographystyle{acl_natbib}

\input{10_appendix}

\end{document}

%% file: 00_abstract.tex
\begin{abstract}
Recent works in dialogue state tracking (DST) focus on an open vocabulary-based setting to resolve scalability and generalization issues of the predefined ontology-based approaches.
However, they are inefficient in that they predict the dialogue state at every turn from scratch.
Here, we consider dialogue state as an explicit fixed-sized memory and propose a selectively overwriting mechanism for more efficient DST.
This mechanism consists of two steps: (1) predicting state operation on each of the memory slots, and (2) overwriting the memory with new values, of which only a few are generated according to the predicted state operations.
Our method decomposes DST into two sub-tasks and guides the decoder to focus only on one of the tasks, thus reducing the burden of the decoder. This enhances the effectiveness of training and DST performance.
Our \ours{} (Selectively Overwriting Memory for Dialogue State Tracking) model achieves state-of-the-art joint goal accuracy with \sotaa{} in MultiWOZ 2.0 and \sotab{} in MultiWOZ 2.1 in an open vocabulary-based DST setting.
In addition, we analyze the accuracy gaps between the current and the ground truth-given situations and suggest that it is a promising direction to improve state operation prediction to boost the DST performance.\footnote{The code is available at \href{https://github.com/clovaai/som-dst}{github.com/clovaai/som-dst}.}


\end{abstract}

%% file: 01_introduction.tex
\section{Introduction}

\begin{figure}[t] 
\centering
\includegraphics[width=0.5\textwidth]{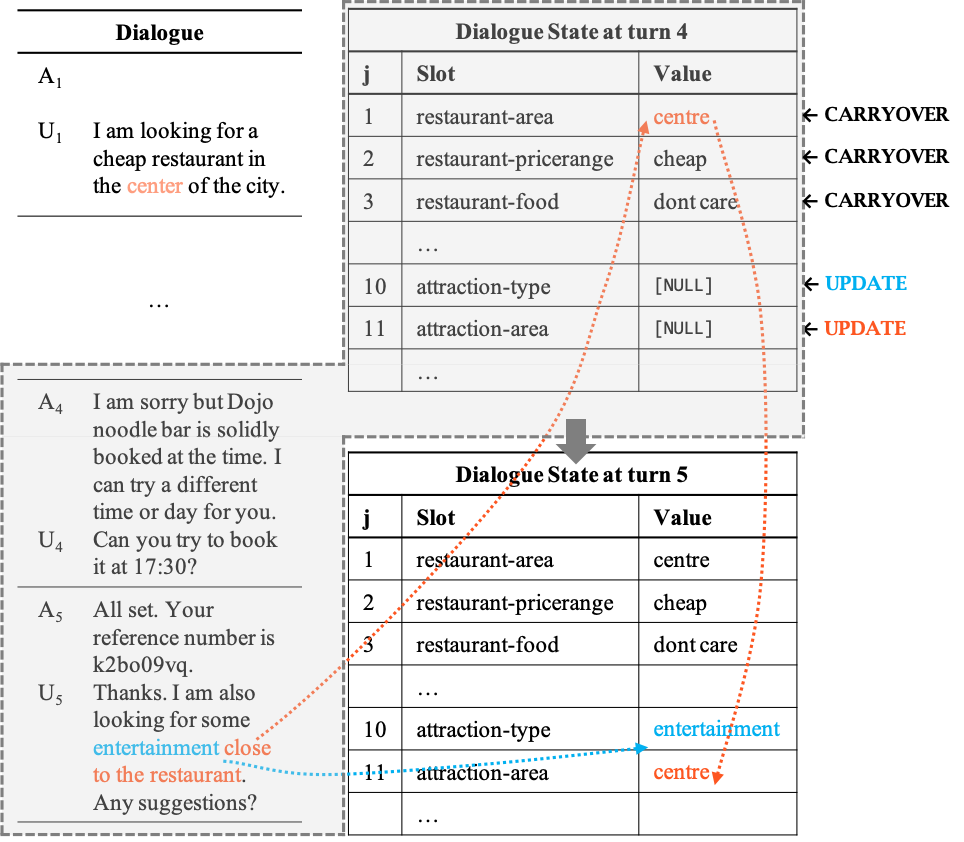}
\caption{
An example of how \ours{} performs dialogue state tracking at a specific dialogue turn (in this case, fifth). The shaded part is the input to the model, and \enquote{Dialogue State at turn 5} at the right-bottom part is the output of the model. Here, \operation{update} operation needs to be performed on the 10th and 11th slot. DST at this turn is challenging since the model requires reasoning over the long-past conversation. However, \ours{} can still robustly perform DST because the previous dialogue state is directly utilized like a memory.
}
\label{fig:opr2}
\end{figure}

Building robust task-oriented dialogue systems has gained increasing popularity in both the research and industry communities~\citep{chen2017survey}.
Dialogue state tracking (DST), one of the essential tasks in task-oriented dialogue systems ~\citep{zhong2018global}, is keeping track of user goals or intentions throughout a dialogue in the form of a set of slot-value pairs, \ie{}, dialogue state.
Because the next dialogue system action is selected based on the current dialogue state, an accurate prediction of the dialogue state has significant importance.

Traditional neural DST approaches assume that all candidate slot-value pairs are given in advance, \ie{}, they perform predefined ontology-based DST \citep{mrkvsic2017neural, zhong2018global, nouri2018toward, lee2019sumbt}.
Most previous works that take this approach perform DST by scoring all possible slot-value pairs in the ontology and selecting the value with the highest score as the predicted value of a slot.
Such an approach has been widely applied to datasets like DSTC2 and WOZ2.0, which have a relatively small ontology size.~\citep{henderson2014second, wen2017network}
Although this approach simplifies the task, it has inherent limitations: (1) it is often difficult to obtain the ontology in advance, especially in a real scenario ~\citep{xu2018end}, (2) predefined ontology-based DST cannot handle previously unseen slot values, and (3) the approach does not scale large since it has to go over all slot-value candidates at every turn to predict the current dialogue state.
Indeed, recent DST datasets often have a large size of ontology; \eg{}, the total number of slot-value candidates in MultiWOZ 2.1 is 4510, while the numbers are much smaller in DSTC2 and WOZ2.0 as 212 and 99, respectively ~\citep{budzianowski2018multiwoz2.0}.

To address these issues, recent methods employ an approach that either directly generates or extracts a value from the dialogue context for every slot, allowing open vocabulary-based DST ~\citep{lei-etal-2018-sequicity, gao2019dialog, wu2019transferable, ren2019comer}.
While this formulation is relatively more scalable and robust to handling unseen slot values, many of the previous works do not efficiently perform DST since they predict the dialogue state from scratch at every dialogue turn.

In this work, we focus on an open vocabulary-based setting and propose \ours{} (\Ours{}).
Regarding dialogue state as a memory that can be selectively overwritten (Figure \ref{fig:opr2}), 
\ours{} decomposes DST into two sub-tasks: 
(1) state operation prediction, which decides the types of the operations to be performed on each of the memory slots, and (2) slot value generation, which generates the values to be newly written on a subset of the memory slots (Figure \ref{fig:overview}). This decomposition allows our model to efficiently generate the values of only a minimal subset of the slots, while many of the previous works generate or extract the values of all slots at every dialogue turn. Moreover, this decomposition reduces the difficulty of DST in an open-vocabulary based setting by clearly separating the roles of the encoder and the decoder. Our encoder, \ie{}, \sop{}, can focus on selecting the slots to pass to the decoder so that the decoder, \ie{}, \svg{}, can focus only on generating the values of those selected slots.
To the best of our knowledge, our work is the first to propose such a selectively overwritable memory-like perspective and a discrete two-step approach on DST.

Our proposed \ours{} achieves state-of-the-art joint goal accuracy in an open vocabulary-based DST setting on two of the most actively studied datasets: MultiWOZ 2.0 and MultiWOZ 2.1.
Error analysis (Section \ref{sec:error_analysis}) further reveals that improving state operation prediction can significantly boost the final DST accuracy.

In summary, the contributions of our work built on top of a perspective that considers dialogue state tracking as selectively overwriting memory are as follows:
\begin{itemize}
    \setlength\itemsep{0em}
    \item Enabling efficient DST, generating the values of a minimal subset of the slots by utilizing the previous dialogue state at each turn.
    \item Achieving state-of-the-art performance on MultiWOZ 2.0 and MultiWOZ 2.1 in an open vocabulary-based DST setting.
    \item Highlighting the potential of improving the state operating prediction accuracy in our proposed framework.
\end{itemize}

%% file: 02_related_work.tex
\input{figs/overview}

\section{Previous Open Vocabulary-based DST}

Many works on recent task-oriented dialogue datasets with a large scale ontology, such as MultiWOZ 2.0 and MultiWOZ 2.1, solve DST in an open vocabulary-based setting \citep{gao2019dialog, wu2019transferable, ren2019comer, le2020endtoend, le2020nonautoregressive}.

\citet{wu2019transferable} show the potential of applying the encoder-decoder framework~\citep{cho2014learning} to open vocabulary-based DST. However, their method is not computationally efficient because it performs autoregressive generation of the values for all slots at every dialogue turn.

\citet{ren2019comer} tackle the drawback of the model of \citet{wu2019transferable}, that their model generates the values of all slots at every dialogue turn, by using a hierarchical decoder.
In addition, they come up with a new notion dubbed Inference Time Complexity (ITC) to compare the efficiency of different DST models.
ITC is calculated using the number of slots $J$ and the number of corresponding slot values $M$.\footnote{The notations used in the work of \citet{ren2019comer} are $n$ and $m$, respectively.}
Following their work, we also calculate ITC in Appendix~\ref{appendix:itc} for comparison.

\citet{le2020nonautoregressive} introduce another work that tackles the efficiency issue. To maximize the computational efficiency, they use a non-autoregressive decoder to generate the slot values of the current dialogue state at once.
They encode the slot type information together with the dialogue context and the delexicalized dialogue context. They do not use the previous turn dialogue state as the input.

\citet{le2020endtoend} process the dialogue context in both domain-level and slot-level. They make the final representation to generate the values using a late fusion approach. They show that there is a performance gain when the model is jointly trained with response generation. However, they still generate the values of every slot at each turn, like \citet{wu2019transferable}.

\citet{gao2019dialog} formulate DST as a reading comprehension task and propose a model named DST Reader that extracts the values of the slots from the input.
They introduce and show the importance of the concept of a slot carryover module, \ie{}, a component that makes a binary decision whether to carry the value of a slot from the previous turn dialogue state over to the current turn dialogue state. The definition and use of discrete operations in our work is inspired by their work.

\citet{zhang2019dsdst} target the issue of ill-formatted strings that generative models suffer from. 
In order to avoid this issue, they take a hybrid approach.
For the slots they categorize as picklist-based slots, they use a predefined ontology-based approach as in the work of \citet{lee2019sumbt}; for the slots they categorize as span-based slots, they use a span extraction-based method like DST-Reader \citep{gao2019dialog}.
However, their hybrid model shows lower performance than when they use only the picklist-based approach. 
Although their solely picklist-based model achieves state-of-the-art joint accuracy in MultiWOZ 2.1, it is done in a predefined ontology-based setting, and thus cannot avoid the scalability and generalization issues of predefined ontology-based DST.

%% file: figs/overview.tex
\begin{figure*}[t!] 
    \centering
    \includegraphics[width=\textwidth]{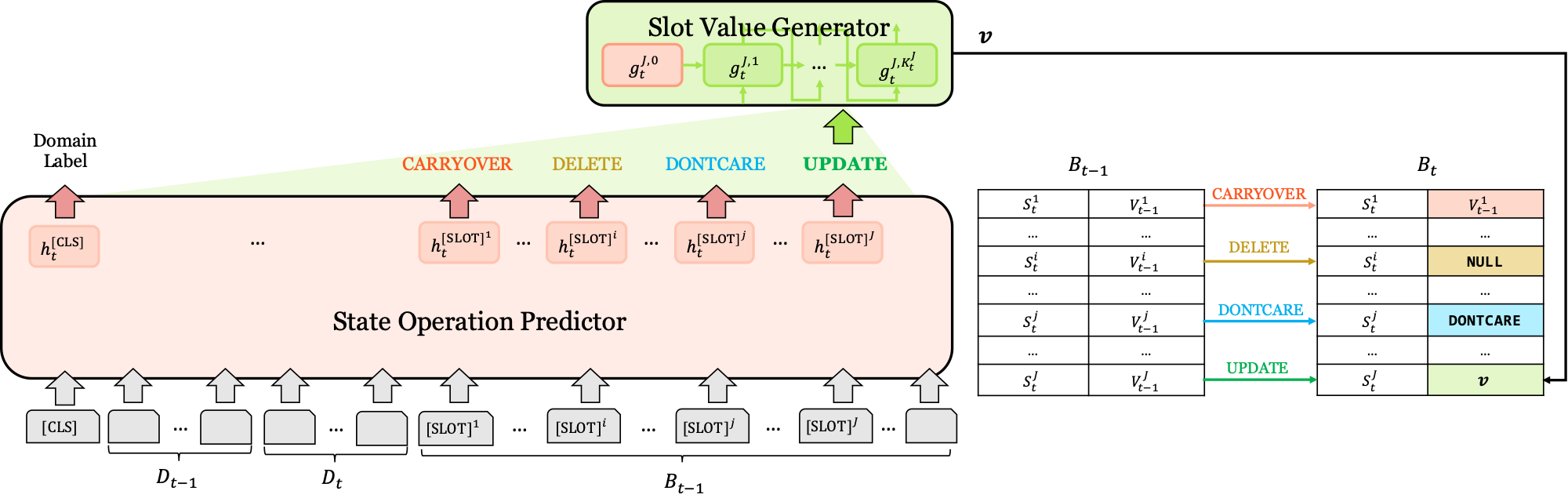}
    \caption{
        The overview of the proposed \ours{}. 
        \ours{} takes the previous turn dialogue utterances $D_{t-1}$, current turn dialogue utterances $D_{t}$, and the previous dialogue state $B_{t-1}$ as the input and outputs the current dialogue state $B_{t}$.
        This is performed by two sub-components: \sop{} and \svg{}.
        \SOP{} takes $D_{t-1}$, $D_{t}$, and $B_{t-1}$ as the input and predicts the operations to perform on each of the slots. Domain classification is jointly performed as an auxiliary task.
        \SVG{} generates the values for the slots that take \operation{update} as the predicted operation.
        The value generation for a slot is done in an autoregressive manner.
    }
    \label{fig:overview}
\end{figure*}

%% file: 03_proposed_method_clean.tex
\section{\Ours{}}
Figure \ref{fig:overview} illustrates the overview of \ours{}.
To describe the proposed \ours{}, we formally define the problem setting in our work.

\medskip \noindent
\textbf{Dialogue State}
We define the dialogue state at turn $t$, $\mathcal{B}_t = \{(S^j, V_t^j) \mid 1 \le j \le J \}$, as a fixed-sized memory whose keys are slots $S^j$ and values are the corresponding slot value $V^j_t$, where $J$ is the total number of such slots.
Following the convention of MultiWOZ 2.0 and MultiWOZ 2.1, we use the term \enquote{slot} to refer to the concatenation of a domain name and a slot name.

\medskip \noindent
\textbf{Special Value}
There are two special values $\texttt{NULL}$ and $\texttt{DONTCARE}$.
$\texttt{NULL}$ means that no information is given about the slot up to the turn.
For instance, the dialogue state before the beginning of any dialogue $\mathcal{B}_0$ has only $\texttt{NULL}$ as the value of all slots.
$\texttt{DONTCARE}$ means that the slot neither needs to be tracked nor considered important in the dialogue at that time.\footnote{Such notions of \enquote{none value} and \enquote{dontcare value} appear in the previous works as well~\cite{wu2019transferable, gao2019dialog,  le2020nonautoregressive, zhang2019dsdst}.}

\medskip \noindent
\textbf{Operation}
At every turn $t$, an operation $r^j_t \in \mathcal{O} = \{\operation{carryover}, \operation{delete}, \operation{dontcare}, \operation{update}\}$ is chosen by the \sop{} (Section \ref{sec:sop}) and performed on each slot $S^j$ to set its current turn corresponding value $V_t^j$. When an operation is performed, it either keeps the slot value unchanged (\operation{carryover}) or changes it to some value different from the previous one (\operation{delete}, \operation{dontcare}, and \operation{update}) as the following.
\begin{equation*}
    {V}_t^j = 
    \begin{cases}
        {V}_{t-1}^j         & \text{if } r^j_t = \operation{carryover} \\
        \texttt{NULL}       & \text{if } r^j_t = \operation{delete} \\
        \texttt{DONTCARE}   & \text{if } r^j_t = \operation{dontcare} \\
        v                   & \text{if } r^j_t = \operation{update} \\
    \end{cases}
\end{equation*}

\noindent{}

The operations that set the value of a slot to a special value (\operation{delete} to \texttt{NULL} and \operation{dontcare} to \texttt{DONTCARE}, respectively) are chosen only when the previous slot value ${V}_{t-1}^j$ is not the corresponding special value. 
\operation{update} operation requires the generation of a new value $v \notin \{{V}_{t-1}^j, \texttt{NULL}, \texttt{DONTCARE}\}$ by \svg{} (Section \ref{sec:svg}). 

State operation predictor performs state operation prediction as a classification task, and \svg{} performs slot value generation to find out the values of the slots on which \operation{update} should be performed.
The two components of \ours{} are jointly trained to predict the current turn dialogue state.

\subsection{State Operation Predictor}
\label{sec:sop}

\textbf{Input Representation}
We denote the representation of the dialogue utterances at turn $t$ as $D_t = A_t \oplus \texttt{;} \oplus U_t \oplus \texttt{[SEP]}$, where $A_t$ is the system response and $U_t$ is the user utterance.
\texttt{;} is a special token used to mark the boundary between $A_t$ and $U_t$, and $\texttt{[SEP]}$ is a special token used to mark the end of a dialogue turn.
We denote the representation of the dialogue state at turn $t$ as $B_t = B_t^1 \oplus \ldots \oplus B_t^J$, where $B_t^j = \texttt{[SLOT]}^j \oplus S^j \oplus \texttt{-} \oplus V^j_t$ is the representation of the $j$-th slot-value pair.
$\texttt{-}$ is a special token used to mark the boundary between a slot and a value.
$\texttt{[SLOT]}^j$ is a special token used to aggregate the information of the $j$-th slot-value pair into a single vector, like the use case of $\texttt{[CLS]}$ token in BERT \cite{devlin2019bert}.
In this work, we use the same special token $\texttt{[SLOT]}$ for all $\texttt{[SLOT]}^j$. 
Our \sop{} employs a pretrained BERT encoder.
The input tokens to the \sop{} are the concatenation of the previous turn dialog utterances, the current turn dialog utterances, and the previous turn dialog state:\footnote{We use only the previous turn dialogue utterances $D_{t-1}$ as the dialogue history, \ie{}, the size of the dialogue history is 1. This is because our model assumes Markov property in dialogues as a part of the input, the previous turn dialogue state $B_{t-1}$, can serve as a compact representation of the whole dialogue history.}
\begin{equation*}
    X_t = \texttt{[CLS]} \oplus D_{t-1} \oplus D_{t} \oplus B_{t-1},
\end{equation*}
\noindent
where \texttt{[CLS]} is a special token added in front of every turn input.
Using the previous dialogue state as the input serves as an explicit, compact, and informative representation of the dialogue history for the model.

When the value of the $j$-th slot at time $t-1$, \ie{}, $V^j_{t-1}$, is $\texttt{NULL}$, we use a special token $\texttt{[NULL]}$ as the input. When the value is $\texttt{DONTCARE}$, we use the string \enquote{dont care} to take advantage of the semantics of the phrase \enquote{don't care} that the pretrained BERT encoder would have already learned.

The input to BERT is the sum of the embeddings of the input tokens $X_t$, segment id embeddings, and position embeddings.
For the segment id, we use 0 for the tokens that belong to $D_{t-1}$ and 1 for the tokens that belong to $D_{t}$ or $B_{t-1}$.
The position embeddings follow the standard choice of BERT.

\medskip \noindent
\textbf{Encoder Output}
The output representation of the encoder is $H_{t} \dimension{|X_t| \times d}$, and $h_{t}^{\texttt{[CLS]}}, h_{t}^{\texttt{[SLOT]}^j} \dimension{d}$ are the outputs that correspond to $\texttt{[CLS]}$ and $\texttt{[SLOT]}^j$, respectively. 
$h_{t}^{X}$, the aggregated sequence representation of the entire input $X_t$, is obtained by a feed-forward layer with a learnable parameter $W_{pool} \dimension{d \times d}$ as:
\begin{equation*}
    h_{t}^{X} = \text{tanh}(W_{pool} \; h_{t}^{\texttt{[CLS]}}).
\end{equation*}

\medskip \noindent
\textbf{State Operation Prediction}
State operation prediction is a four-way classification performed on top of the encoder output for each slot representation $h_{t}^{\texttt{[SLOT]}^j}$:
\begin{equation*}
\begin{aligned}
    P^j_{opr,t} &= \text{softmax}(W_{opr} \; h_{t}^{\texttt{[SLOT]}^j}), \\
\end{aligned}
\end{equation*}
\noindent
where $W_{opr} \dimension{|\mathcal{O}| \times d}$ is a learnable parameter and $P^j_{opr,t} \dimension{|\mathcal{O}|}$ is the probability distribution over operations for the $j$-th slot at turn $t$.
In our formulation, $|\mathcal{O}| = 4$, because $\mathcal{O} = \{\operation{carryover}$, $\operation{delete}$, $\operation{dontcare}$, $\operation{update} \}$.

Then, the operation is determined by $r^j_t = \text{argmax}(P^j_{opr,t})$ and the slot value generation is performed on only the slots whose operation is $\operation{update}$.
We define the set of the slot indices which require the value generation as $\mathbb{U}_t = \{j \mid r^j_t = \operation{update}\}$, and its size as $J^{\prime}_t = |\mathbb{U}_t|$.

\subsection{Slot Value Generator}
\label{sec:svg}
For each $j$-th slot such that $j \in \mathbb{U}_t$, the \svg{} generates a value.
Our \svg{} differs from the generators of many of the previous works because it generates the values for only $J^{\prime}_t$ number of slots, not $J$.
In most cases, $J^{\prime}_t \ll J$, so this setup enables an efficient computation where only a small number of slot values are newly generated.

We use Gated Recurrent Unit (GRU) \citep{cho2014properties} decoder like \citet{wu2019transferable}.
GRU is initialized with $g^{j,0}_{t} = h_{t}^\texttt{X}$ and $e_{t}^{j,0} = h_{t}^{\texttt{[SLOT]}^j}$, and recurrently updates the hidden state $g^{j,k}_{t} \dimension{d}$ by taking a word embedding $e_{t}^{j,k}$ as the input until $\texttt{[EOS]}$ token is generated: 
\begin{equation*}
    g^{j,k}_{t} = \text{GRU}(g^{j,k-1}_{t}, e_{t}^{j,k}).
\end{equation*}
\noindent

The decoder hidden state is transformed to the probability distribution over the vocabulary at the $k$-th decoding step, where $E \dimension{d_{vcb} \times d}$ is the word embedding matrix shared across the encoder and the decoder, such that $d_{vcb}$ is the vocabulary size.
\begin{equation*}
    P^{j,k}_{vcb,t} = \text{softmax}(E \; g^{j,k}_{t}) \dimension{d_{vcb}}.
\end{equation*}

As the work of \citet{wu2019transferable}, we use the soft-gated copy mechanism \cite{see2017get} to get the final output distribution $P^{j,k}_{val,t}$ over the candidate value tokens:
\begin{equation*}
\begin{aligned}
    &P^{j,k}_{ctx,t} = \text{softmax}(H_{t} \; g^{j,k}_{t}) \dimension{|X_t|}, \\
    &P^{j,k}_{val,t} = \alpha P^{j,k}_{vcb,t} + (1 - \alpha) P^{j,k}_{ctx,t},
\end{aligned}
\end{equation*}
\noindent
such that $\alpha$ is a scalar value computed as:
\begin{equation*}
    \alpha = \text{sigmoid}(W_1 \; [g^{j,k}_{t}; e_{t}^{j,k}; c_{t}^{j,k}]),
\end{equation*}
\noindent
where $W_1 \dimension{1 \times (3d)}$ is a learnable parameter and $c_{t}^{j,k} = P^{j,k}_{ctx,t} \; H_{t} \dimension{d}$ is a context vector.

\subsection{Objective Function}
During training, we jointly optimize both \sop{} and \svg{}.

\medskip \noindent
\textbf{\SOP{}}
In addition to the state operation classification, we use domain classification as an auxiliary task to force the model to learn the correlation of slot operations and domain transitions in between dialogue turns.
Domain classification is done with a softmax layer on top of $h_{t}^{X}$:
\begin{equation*}
    P_{dom,t} = \text{softmax}(W_{dom} \; h_{t}^X),
\end{equation*}
\noindent
where $W_{dom} \dimension{d_{dom} \times d}$ is a learnable parameter and $P_{dom,t} \dimension{d_{dom}}$ is the probability distribution over domains at turn $t$.
$d_{dom}$ is the number of domains defined in the dataset.

The loss for each of state operation classification and domain classification is the average of the negative log-likelihood, as follows:
\begin{equation*}
\begin{aligned}
    &L_{opr,t} = -\frac{1}{J} \sum_{j=1}^{J} (Y_{opr,t}^{j})^\intercal \log(P_{opr,t}^j), \\
    &L_{dom,t} = -(Y_{dom,t})^\intercal \log(P_{dom,t}), \\
\end{aligned}
\end{equation*}

\noindent
where $Y_{dom,t} \dimension{d_{dom}}$ is the one-hot vector for the ground truth domain and $Y_{opr,t}^j \dimension{|\mathcal{O}|}$ is the one-hot vector for the ground truth operation for the $j$-th slot.

\medskip \noindent
\textbf{\SVG{}}
The objective function to train \svg{} is also the average of the negative log-likelihood:
\begin{equation*}
\begin{split}
    L_{svg,t}
    = -\frac{1}{|\mathbb{U}_t|} \sum_{j \in \mathbb{U}_t} \bigg[ \frac{1}{K^j_t} \sum_{k=1}^{K^j_t} (Y_{val,t}^{j,k})^\intercal \log(P_{val,t}^{j,k}) \bigg],
\end{split}
\end{equation*}

\noindent
where $K^j_t$ is the number of tokens of the ground truth value that needs to be generated for the $j$-th slot. $Y_{val,t}^{j,k} \dimension{d_{vcb}}$ is the one-hot vector for the ground truth token that needs to be generated for the $j$-th slot at the $k$-th decoding step.

Therefore, the final joint loss $L_{joint,t}$ to be minimized at dialogue turn $t$ is the sum of the losses mentioned above:
\begin{equation*}
    L_{joint,t} = L_{opr,t} + L_{dom, t} + L_{svg,t}.
\end{equation*}

%% file: 04_experimental_setup.tex
\section{Experimental Setup}

\subsection{Datasets}

%
%
%

We use MultiWOZ 2.0 ~\citep{budzianowski2018multiwoz2.0} and MultiWOZ 2.1 ~\citep{eric2019multiwoz2.1} as the datasets in our experiments. These datasets are two of the largest publicly available multi-domain task-oriented dialogue datasets, including about 10,000 dialogues within seven domains. 
MultiWOZ 2.1 is a refined version of MultiWOZ 2.0 in which the annotation errors are corrected.\footnote{See Table \ref{table:data-statistics} in Appendix \ref{appendix:data-stat} for more details of MultiWOZ 2.1.}

Following ~\citet{wu2019transferable}, we use only five domains (\textit{restaurant}, \textit{train}, \textit{hotel}, \textit{taxi}, \textit{attraction}) excluding \textit{hospital} and \textit{police}.\footnote{The excluded domains take up only a small portion of the dataset and do not even appear in the test set.} Therefore, the number of domains $d_{dom}$ is 5 and the number of slots $J$ is 30 in our experiments. We use the script provided by ~\citet{wu2019transferable} to preprocess the datasets.\footnote{\href{https://github.com/jasonwu0731/trade-dst}{github.com/jasonwu0731/trade-dst}}

\subsection{Training}
We employ the pretrained  BERT-base-uncased model\footnote{\href{https://github.com/huggingface/transformers}{github.com/huggingface/transformers}} for \sop{} and one GRU ~\citep{cho2014properties} for \svg{}. The hidden size of the decoder is the same as that of the encoder, $d$, which is 768. The token embedding matrix of \svg{} is shared with that of \sop{}. We use BertAdam as our optimizer~\citep{kingma2015adam}.
We use greedy decoding for \svg{}.

The encoder of \sop{} makes use of a pretrained model, whereas the decoder of \svg{} needs to be trained from scratch. Therefore, we use different learning rate schemes for the encoder and the decoder. We set the peak learning rate and warmup proportion to 4e-5 and 0.1 for the encoder and 1e-4 and 0.1 for the decoder, respectively.
We use a batch size of 32 and set the dropout \cite{srivastava2014dropout} rate to 0.1. We also utilize word dropout~\cite{bowman-etal-2016-generating} by randomly replacing the input tokens with the special \texttt{[UNK]} token with the probability of 0.1. The max sequence length for all inputs is fixed to 256.

We train \sop{} and \svg{} jointly for 30 epochs and choose the model that reports the best performance on the validation set.
During training, we use the ground truth state operations and the ground truth previous turn dialogue state instead of the predicted ones.
When the dialogue state is fed to the model, we randomly shuffle the slot order with a rate of 0.5.
This is to make \sop{} exploit the semantics of the slot names and not rely on the position of the slot tokens or a specific slot order.
During inference or when the slot order is not shuffled, the slots are sorted alphabetically.
We use teacher forcing 50\% of the time to train the decoder.

All experiments are performed on NAVER Smart Machine Learning (NSML) platform~\cite{sung2017nsml,kim2018nsml}.
All the reported results of \ours{} are averages over ten runs.

\subsection{Baseline Models}
We compare the performance of \ours{} with both predefined ontology-based models and open vocabulary-based models.

\medskip \noindent
\textbf{FJST} uses a bidirectional LSTM to encode the dialogue history and uses a feed-forward network to predict the value of each slot
~\citep{eric2019multiwoz2.1}.

\medskip \noindent
\textbf{HJST} is proposed together with FJST; it encodes the dialogue history using an LSTM like FJST but uses a hierarchical network ~\citep{eric2019multiwoz2.1}.

\medskip \noindent
\textbf{SUMBT} exploits BERT-base as the encoder for the dialogue context and slot-value pairs. After encoding them, it scores every candidate slot-value pair in a non-parametric manner using a distance measure~\citep{lee2019sumbt}.

\medskip \noindent
\textbf{HyST} employs a hierarchical RNN encoder and takes a hybrid approach that incorporates both a predefined ontology-based setting and an open vocabulary-based setting~\citep{goel2019hyst}.

\medskip \noindent
\textbf{DST Reader} formulates the problem of DST as an extractive QA task; it uses BERT-base to make the contextual word embeddings and extracts the value of the slots from the input as a span~\citep{gao2019dialog}.

\medskip \noindent
\textbf{TRADE} encodes the whole dialogue context with a bidirectional GRU and decodes the value for every slot using a copy-augmented GRU decoder~\citep{wu2019transferable}.

\medskip \noindent
\textbf{COMER} uses BERT-large as a feature extractor and a hierarchical LSTM decoder to generate the current turn dialogue state itself as the target sequence~\citep{ren2019comer}.

\medskip \noindent
\textbf{NADST} uses a Transformer-based non-autoregressive decoder to generate the current turn dialogue state~\citep{le2020nonautoregressive}.

\medskip \noindent
\textbf{ML-BST} uses a Transformer-based architecture to encode the dialogue context with the domain and slot information and combines the outputs in a late fusion approach.
Then, it generates the slot values and the system response jointly~\citep{le2020endtoend}.

\medskip \noindent
\textbf{DS-DST} uses two BERT-base encoders and takes a hybrid approach of predefined ontology-based DST and open vocabulary-based DST. It defines picklist-based slots for classification similarly to SUMBT and span-based slots for span extraction like DST Reader~\citep{zhang2019dsdst}.

\medskip \noindent
\textbf{DST-picklist} is proposed together with DS-DST and uses a similar architecture, but it performs only predefined ontology-based DST considering all slots as picklist-based slots~\citep{zhang2019dsdst}.

%% file: 05_experimental_results.tex
\section{Experimental Results}
\label{sec:exp_results}

\subsection{Joint Goal Accuracy}
\input{tabs/joint_goal_accuracy}

Table \ref{table:main-results} shows the joint goal accuracy of \ours{} and other models on the test set of MultiWOZ 2.0 and MultiWOZ 2.1. Joint goal accuracy is an accuracy which checks whether all slot values predicted at a turn exactly match the ground truth values.

As shown in the table, \ours{} achieves state-of-the-art performance in an open vocabulary-based setting. Interestingly, on the contrary to the previous works, our model achieves higher performance on MultiWOZ 2.1 than on MultiWOZ 2.0.
This is presumably because our model, which explicitly uses the dialogue state labels as input, benefits more from the error correction on the state annotations done in MultiWOZ 2.1.\footnote{\citet{eric2019multiwoz2.1} report that the correction of the annotations done in MultiWOZ 2.1 changes about 32\% of the state annotations of MultiWOZ 2.0, which indicates that MultiWOZ 2.0 consists of many annotation errors.}
\pagebreak 


\subsection{Domain-Specific Accuracy}
\input{tabs/domain_results}

Table \ref{table:domain-results} shows the domain-specific results of our model and the concurrent works which report such results ~\citep{le2020endtoend, le2020nonautoregressive}. Domain-specific accuracy is the accuracy measured on a subset of the predicted dialogue state, where the subset consists of the slots specific to a domain.

While the performance is similar to or a little lower than that of other models in other domains, \ours{} outperforms other models in \textit{taxi} and \textit{train} domains. This implies that the state-of-the-art joint goal accuracy of our model on the test set comes mainly from these two domains.

A characteristic of the data from these domains is that they consist of challenging conversations; the slots of these domains are filled with more diverse values than other domains,\footnote{The statistics of the slot value vocabulary size are shown in Table \ref{table:slot-value-vocab-size} in Appendix \ref{appendix:data-stat}.} and there are more than one domain changes, \ie{}, the user changes the conversation topic during a dialogue more than once. For a specific example, among the dialogues where the domain switches more than once, the number of conversations that end in \textit{taxi} domain is ten times more than in other cases. A more detailed statistics are given in Table \ref{table:transition-stats} in Appendix \ref{appendix:data-stat}.

Therefore, we assume our model performs relatively more robust DST in such challenging conversations. 
We conjecture that this strength attributes to the effective utilization of the previous turn dialogue state in its explicit form, like using a memory; the model can explicitly keep even the information mentioned near the beginning of the conversation and directly copy the values from this memory whenever necessary. Figure \ref{fig:opr2} shows an example of a complicated conversation in MultiWOZ 2.1, where our model accurately predicts the dialogue state. More sample outputs of \ours{} are provided in Appendix \ref{appendix:examples}.

%
%
%
%

%% file: tabs/joint_goal_accuracy.tex
\begin{table}[t!]
    \centering
    \caption{
        Joint goal accuracy on the test set of MultiWOZ 2.0 and 2.1. * indicates a result borrowed from \citet{eric2019multiwoz2.1}.
        HyST and DS-DST use a hybrid approach, partially taking advantage of the predefined ontology.
        $\dagger$ indicates the case where BERT-large is used for our model.
    }
    \label{table:main-results}
    \footnotesize
    \begin{threeparttable}
    \begin{tabular*}{\columnwidth}{lrr}
        \toprule
             & {\scriptsize \makecell[r]{MultiWOZ \\ 2.0}} &  {\scriptsize \makecell[r]{ MultiWOZ \\ 2.1}} \\
        \midrule
        \textbf{Predefined Ontology} \\
        \midrule
            \quad HJST$^*$ \citep{eric2019multiwoz2.1} & 38.40 & 35.55 \\
            \quad FJST$^*$ \citep{eric2019multiwoz2.1} & 40.20 & 38.00 \\
            \quad SUMBT \citep{lee2019sumbt} & 42.40 & - \\
            \quad HyST$^*$ \citep{goel2019hyst} & 42.33 & 38.10 \\
            \quad DS-DST \citep{zhang2019dsdst} & - & 51.21 \\
            \quad DST-picklist \citep{zhang2019dsdst} & - & \textbf{53.30} \\
        \midrule
        \textbf{Open Vocabulary} \\
        \midrule
            \quad DST Reader$^*$ \citep{gao2019dialog} & 39.41 & 36.40 \\
            \quad TRADE$^*$ ~\citep{wu2019transferable} & 48.60 & 45.60 \\
            \quad COMER \citep{ren2019comer} & 48.79 & - \\
            \quad NADST \citep{le2020nonautoregressive} & 50.52 & 49.04 \\
            \quad ML-BST \citep{le2020endtoend} & - & 50.91 \\
            \quad \ours{} (ours) & \textbf{51.72} & \textbf{53.01} \\
        \midrule
            \quad \ours{}$^\dagger$ (ours) & \textbf{52.32} & \textbf{53.68} \\
        \bottomrule

    \end{tabular*}
    \end{threeparttable}
\end{table}

%% file: tabs/domain_results.tex
\begin{table}[t!]
    \centering
    \caption{Domain-specific results on the test set of MultiWOZ 2.1.
    Our model outperforms other models in \textit{taxi} and \textit{train} domains.
    }
    \label{table:domain-results}
    \footnotesize
    \begin{threeparttable}
    \begin{tabular*}{\columnwidth}{llr@{\extracolsep{\fill}}r}
        \toprule
        Domain & Model & \makecell[r]{Joint \\ Accuracy} & \makecell[r]{Slot \\ Accuracy} \\
        \midrule
        Attraction & NADST & 66.83 & 98.79 \\
            & ML-BST & \textbf{70.78} & 99.06 \\
            & \ours{} (ours) & 69.83 & 98.86 \\
        \cmidrule{1-4}
        Hotel & NADST & 48.76 & 97.70 \\
            & ML-BST & 49.52 & 97.50 \\
            & \ours{} (ours) & \textbf{49.53} & 97.35 \\
        \cmidrule{1-4}
        Restaurant & NADST & 65.37 & 98.78 \\
            & ML-BST & \textbf{66.50} & 98.76 \\
            & \ours{} (ours) & 65.72 & 98.56 \\
        \cmidrule{1-4}
        Taxi & NADST & 33.80 & 96.69 \\
            & ML-BST & 23.05 & 96.42 \\
            & \ours{} (ours) & \textbf{59.96} & 98.01 \\
        \cmidrule{1-4}
        Train & NADST & 62.36 & 98.36 \\
            & ML-BST & 65.12 & 90.22 \\
            & \ours{} (ours) & \textbf{70.36} & 98.67 \\
        \bottomrule
    \end{tabular*}
    \end{threeparttable}
\end{table}

%% file: 06_analysis.tex
\section{Analysis}
\subsection{Choice of State Operations}
\label{appendix:ablation-study}
\noindent
Table \ref{table:ablation} shows the joint goal accuracy where the four-way state operation prediction changes to two-way, three-way, or six-way.

The joint goal accuracy drops when we use two-way state operation prediction, which is a binary classification of whether to (1) carry over the previous slot value to the current turn or (2) generate a new value, like \citet{gao2019dialog}.
We assume the reason is that it is better to separately model operations \operation{delete}, \operation{dontcare}, and \operation{update} that correspond to the latter class of the binary classification, since the values of \operation{delete} and \operation{dontcare} tend to appear implicitly while the values for \operation{update} are often explicitly expressed in the dialogue.

We also investigate the performance when only three operations are used or two more state operations, \operation{yes} and \operation{no}, are used. \operation{yes} and \operation{no} represent the cases where yes or no should be filled as the slot value, respectively.
The performance drops in all of the cases.

\input{tabs/ablation}

\subsection{Error Analysis}
\label{sec:error_analysis}

\input{tabs/error_analysis}

Table~\ref{table:error-analysis} shows the joint goal accuracy of the combinations of the cases where the ground truth is used or not for each of the previous turn dialogue state, state operations at the current turn, and slot values for \operation{update} at the current turn.
From this result, we analyze which of \sop{} and \svg{} is more responsible for the error in the joint goal prediction, under the cases where error propagation occurs or not.

Among the absolute error of 46.99\% made under the situation that error propagation occurs, \ie{}, the dialogue state predicted at the previous turn is fed to the model, it could be argued that 92.85\% comes from \sop{}, 21.6\% comes from \svg{}, and 14.45\% comes from both of the components.
This indicates that at least 78.4\% to 92.85\% of the error comes from \sop{}, and at least 7.15\% to 21.6\% of the error comes from \svg{}.
\footnote{The calculation of the numbers in the paragraph is done as follows. (The figures in the paragraph immediately below are calculated in the same way.)\\

\scriptsize
\begin{tabular}{@{}ll} 
$100 - 53.01 = 46.99$ & $92.85 + 21.6 - 100 = 14.45$ \\
$(100 - 56.37) / 46.99 = 92.85$ & $92.85 - 14.45 = 78.4$ \\
$(100 - 89.85) / 46.99 = 21.6$ & $21.6 - 14.45 = 7.15$ \\
\end{tabular}


}

Among the absolute error of 19\% made under the error propagation-free situation, \ie{}, ground truth previous turn dialogue state is fed to the model, it could be argued that 90.53\% comes from \sop{}, 19.63\% comes from \svg{}, and 10.16\% comes from both of the components.
This indicates that at least 80.37\% to 90.53\% of the error comes from \sop{}, and at least 9.47\% to 19.63\% of the error comes from \svg{}.

\input{tabs/state_operation}

Error propagation that comes from using the dialogue state predicted at the previous turn increases the error 2.47 (=$\frac{100-53.01}{100-81.00}$) times. Both with and without error propagation, a relatively large amount of error comes from \sop{}, implying that a large room for improvement currently exists in this component.
Improving the state operation prediction accuracy, e.g., by tackling the class imbalance shown in Table \ref{table:state_operation}, may have the potential to increase the overall DST performance by a large margin.

\subsection{Efficiency Analysis}
\label{sec:efficiency_analysis}
\input{tabs/efficiency_analysis.tex}


In Table \ref{table:efficiency}, we compare the number of slot values generated at a turn among various open vocabulary-based DST models that use an autoregressive decoder.

The maximum number of slots whose values are generated by our model at a turn, \ie{}, the number of slots on which \operation{update} should be performed, is 9 at maximum and only 1.14 on average in the test set of MultiWOZ 2.1.

On the other hand, TRADE and ML-BST generate the values of all the 30 slots at every turn of a dialogue. COMER generates only a subset of the slot values like our model, but it generates the values of all the slots that have a non-\texttt{NULL} value at a turn, which is 18 at maximum and 5.72 on average.

Table~\ref{table:latency} shows the latency of \ours{} and several other models.
We measure the inference time for a dialogue turn of MultiWOZ 2.1 on Tesla V100 with a batch size of 1.
The models used for comparison are those with official public implementations.

It is notable that the inference time of \ours{} is about 12.5 times faster than TRADE, which consists of only two GRUs. Moreover, the latency of \ours{} is compatible with that of NADST, which explicitly uses non-autoregressive decoding, while \ours{} achieves much higher joint goal accuracy.
This shows the efficiency of the proposed selectively overwriting mechanism of \ours{}, which generates only the minimal slot values at a turn.

In Appendix~\ref{appendix:itc}, we also investigate Inference Time Complexity (ITC) proposed in the work of \citet{ren2019comer}, which defines the efficiency of a DST model using $J$, the number of slots, and $M$, the number of values of a slot.



%% file: tabs/ablation.tex
\begin{table}[!t]
    \centering
    \label{table:ablation-study}
    \caption{
        Joint goal accuracy on the MultiWOZ 2.1 test set when the four-way state operation prediction changes to two-way, three-way, or six-way.
    }
    \label{table:ablation}
    \footnotesize
    \begin{threeparttable}
    \begin{tabular*}{\columnwidth}{cl@{\extracolsep{\fill}}r}
        \toprule
        \multicolumn{2}{c}{\multirow{2}{*}{State Operations}} & Joint \\
        & & Accuracy \\
        \midrule
        \multirow{2}{*}{4}
        & \operation{carryover}, \operation{delete}, 
        & \multirow{2}{*}{53.01} \\
        & \operation{dontcare}, \operation{update} & \\
        \midrule
        2 
        & \operation{carryover}, \operation{non-carryover} & 52.06 \\
        3 
        & \operation{carryover}, \operation{dontcare}, \operation{update} & 52.63 \\        
        3 
        & \operation{carryover}, \operation{delete}, \operation{update} & 52.64 \\
        
        \multirow{2}{*}{6} 
        & \operation{carryover}, \operation{delete},  
        & \multirow{2}{*}{52.97} \\
        
        & \operation{dontcare}, \operation{update}, \operation{yes}, \operation{no} & \\
        
        \bottomrule
    \end{tabular*}
    \end{threeparttable}
\end{table}

%% file: tabs/error_analysis.tex
\begin{table}[t!]
    \centering
    \caption{
        Joint goal accuracy of the current and the ground truth-given situations.
        Relative error rate is the proportion of the error when 100\% is set as the error where no ground truth is used for SOP and SVG.
        (GT: Ground Truth, SOP: State Operation Prediction, SVG: Slot Value Generation, Pred: Predicted)
    }
    \label{table:error-analysis}
    \footnotesize
    \begin{threeparttable}
    \begin{tabular*}{\columnwidth}{cc@{\extracolsep{\fill}}crr}
        \toprule
            &  GT &  GT  & Joint & Relative \\
            & SOP & SVG & Accuracy & Error Rate \\
        \midrule
        \multirow{4}{*}{\makecell[c]{Pred $B_{t-1}$ \\ (w/ Error \\ Propagation)}} & & & 53.01 & 100.0 \\
            & & \checkmark & 56.37 & 92.85 \\
            & \checkmark & & 89.85 & 21.60 \\
            & \checkmark & \checkmark & 100.0 & 0.00 \\
        \midrule
        \multirow{4}{*}{\makecell[c]{GT $B_{t-1}$ \\ (w/o Error \\ Propagation)}} & & & 81.00 & 100.0 \\
            & & \checkmark & 82.80 & 90.53 \\
            & \checkmark & & 96.27 & 19.63  \\
            & \checkmark & \checkmark & 100.0 & 0.00 \\
        \bottomrule

    \end{tabular*}
    \end{threeparttable}
\end{table}

%% file: tabs/state_operation.tex
\begin{table}[!t]
    \setlength{\tabcolsep}{3.5pt}
    \centering
    \caption{Statistics of the number of state operations and the corresponding F1 scores of our model in MultiWOZ 2.1.}
    \label{table:state_operation}
    \footnotesize
    \begin{threeparttable}
    \begin{tabular*}{\columnwidth}{l@{\extracolsep{\fill}}rrrr}
        \toprule
            & \multicolumn{3}{c}{\# Operations} & F1 score\\
        \cmidrule{2-4}
        Operation Type & Train & Valid & Test & Test\\
        \midrule
        \operation{carryover} & 1,584,757 & 212,608 & 212,297 & 98.66 \\
        \operation{update} & 61,628 & 8,287 & 8,399 & 80.10 \\
        \operation{dontcare} & 1,911 & 155 & 235 & 32.51 \\
        \operation{delete} & 1,224 & 80 & 109 & 2.86 \\
        \bottomrule
    \end{tabular*}
    \end{threeparttable}
\end{table}
\textcolor{white}{.}

%% file: tabs/efficiency_analysis.tex

\begin{table}[t!]
    \setlength{\tabcolsep}{3.5pt}
    \centering
    \caption{The minimum, average, and maximum number of slots whose values are generated at a turn, calculated on the test set of MultiWOZ 2.1.}
    \label{table:efficiency}
    \footnotesize
    \begin{threeparttable}
    \begin{tabular*}{\columnwidth}{l@{\extracolsep{\fill}}rrr}
        \toprule
        Model & Min \# & Avg \# & Max \# \\
        \midrule
        TRADE & 30 & 30 & 30 \\
        ML-BST & 30 & 30 & 30 \\
        COMER & 0 & 5.72 & 18 \\
        \ours{} (ours) & 0 & 1.14 & 9 \\
        \bottomrule
    \end{tabular*}
    \end{threeparttable}
\end{table}

\begin{table}[t!]
    \setlength{\tabcolsep}{3.5pt}
    \centering
    \caption{Average inference time per dialogue turn of MultiWOZ 2.1 test set, measured on Tesla V100 with a batch size of 1. $\dagger$ indicates the case where BERT-large is used for our model.}
    \label{table:latency}
    \footnotesize
    \begin{threeparttable}
    \begin{tabular*}{\columnwidth}{l@{\extracolsep{\fill}}rr}
        \toprule
        Model & Joint Accuracy & Latency \\
        \midrule
        TRADE & 45.60 & 340 ms  \\
        NADST & 49.04 & 26 ms  \\
        \ours{} (ours) & \textbf{53.01} & 27 ms  \\
        \ours{}$^\dagger$ (ours) & \textbf{53.68} & 40 ms \\
        \bottomrule
    \end{tabular*}
    \end{threeparttable}
\end{table}

%% file: 07_conclusion.tex
\section{Conclusion}
We propose \ours{}, an open vocabulary-based dialogue state tracker that regards dialogue state as an explicit memory that can be selectively overwritten. \ours{} decomposes dialogue state tracking into state operation prediction and slot value generation. This setup makes the generation process efficient because the values of only a minimal subset of the slots are generated at each dialogue turn. \ours{} achieves state-of-the-art joint goal accuracy on both MultiWOZ 2.0 and MultiWOZ 2.1 datasets in an open vocabulary-based setting. \ours{} effectively makes use of the explicit dialogue state and discrete operations to perform relatively robust DST even in complicated conversations.
Further analysis shows that improving state operation prediction has the potential to increase the overall DST performance dramatically.
From this result, we propose that tackling DST with our proposed problem definition is a promising future research direction.

%% file: 08_acknowledgements.tex
\section*{Acknowledgments}
The authors would like to thank the members of Clova AI for proofreading this manuscript.

%% file: 10_appendix.tex
\appendix

\onecolumn
\begin{appendices}

\section{Data Statistics}
\label{appendix:data-stat}

\begin{table}[!ht]
    \centering
    \caption{Data Statistics of MultiWOZ 2.1.}
    \label{table:data-statistics}
    \footnotesize
    \begin{threeparttable}
    \begin{tabularx}{\columnwidth}{llrrrrrr}
        \toprule
            & & \multicolumn{3}{c}{\# of Dialogues} & \multicolumn{3}{c}{\# of Turns} \\
        \cmidrule{3-8}
        Domain    & Slots & Train & Valid & Test & Train & Valid & Test \\
        \midrule
        Attraction & \makecell[Xt]{area, name, type} 
        & \quad 2,717 & 401 & 395 
        & \quad 8,073 & 1,220 & 1,256 \\
        \cmidrule{1-8}
        Hotel & \makecell[Xt]{price range, type, parking, book stay, book day, book people, area, stars, internet, name} 
        & \quad 3,381 & 416 & 394
        & \quad 14,793 & 1,781 & 1,756 \\
        \cmidrule{1-8}
        Restaurant & \makecell[Xt]{food, price range, area, name, book time, book day, book people} 
        & \quad 3,813 & 438 & 437
        & \quad 15,367 & 1,708 & 1,726 \\
        \cmidrule{1-8}
        Taxi & \makecell[Xt]{leave at, destination, departure, arrive by} 
        & \quad 1,654 & 207 & 195 
        & \quad 4,618 & 690 & 654 \\
        \cmidrule{1-8}
        Train & \makecell[Xt]{destination, day, departure, arrive by, book people, leave at}
        & \quad 3,103 & 484 & 494
        & \quad 12,133 & 1,972 & 1,976 \\
        \bottomrule
    \end{tabularx}
    \end{threeparttable}
\end{table}


\newcolumntype{A}[1]{>{\raggedright\let\newline\\\arraybackslash\hspace{0pt}}m{#1}}
\newcolumntype{B}[1]{>{\raggedleft\let\newline\\\arraybackslash\hspace{0pt}}m{#1}}

\begin{table}[!h]
    \setlength{\tabcolsep}{3.5pt}
    \centering
    \caption{Statistics of the slot value vocabulary size in MultiWOZ 2.1.}
    \label{table:slot-value-vocab-size}
    \footnotesize
    \begin{threeparttable}
    \begin{tabular*}{\columnwidth}{A{7.3cm}B{0.7cm}B{3.5cm}B{3.5cm}}
        \toprule
            & \multicolumn{3}{c}{Slot Value Vocabulary Size}\\
            \cmidrule{2-4}
        Slot Name & Train & Valid  & Test \\
        \midrule
taxi-destination & 373 & 213 & 213 \\
taxi-departure & 357 & 214 & 203 \\
restaurant-name & 202 & 162 & 162 \\
attraction-name & 186 & 145 & 149 \\
train-leaveat & 146 & 69 & 117 \\
train-arriveby & 112 & 64 & 101 \\
restaurant-food & 111 & 81 & 70 \\
taxi-leaveat & 105 & 68 & 65 \\
hotel-name & 93 & 65 & 58 \\
restaurant-book time & 64 & 50 & 51 \\
taxi-arriveby & 95 & 49 & 46 \\
train-destination & 27 & 25 & 24 \\
train-departure & 34 & 23 & 23 \\
attraction-type & 31 & 17 & 17 \\
train-book people & 11 & 9 & 9 \\
hotel-book people & 8 & 8 & 8 \\
restaurant-book people & 9 & 8 & 8 \\
hotel-book day & 13 & 7 & 7 \\
hotel-stars & 9 & 7 & 7 \\
restaurant-book day & 10 & 7 & 7 \\
train-day & 8 & 7 & 7 \\
attraction-area & 7 & 6 & 6 \\
hotel-area & 7 & 6 & 6 \\
restaurant-area & 7 & 6 & 6 \\
hotel-book stay & 10 & 5 & 5 \\
hotel-parking & 4 & 4 & 4 \\
hotel-pricerange & 7 & 5 & 4 \\
hotel-type & 5 & 5 & 4 \\
restaurant-pricerange & 5 & 4 & 4 \\
hotel-internet & 3 & 3 & 3 \\
        \bottomrule
    \end{tabular*}
    \end{threeparttable}
\end{table}

\begin{table}[!h]
    \caption{Statistics of domain transition in the test set of MultiWOZ 2.1. There are 140 dialogues with more than one domain transition that end with \textit{taxi} domain. The cases where domain switches more than once and ends in \textit{taxi} are shown in bold. The total number of dialogues with more than one domain transition is 175. We can view these as complicated dialogues.}
    \label{table:transition-stats}
    \centering
    \footnotesize
    \begin{threeparttable}
    \newcolumntype{L}{>{\raggedright\arraybackslash}X}
    \newcolumntype{C}{>{\centering\arraybackslash}X}
    \newcolumntype{R}{>{\raggedleft\arraybackslash}X}
    \begin{tabularx}{\textwidth}{LLLLR}
        \toprule
        \multicolumn{4}{c}{Domain Transition} & \\
        \cmidrule{1-4}
        First & Second & Third & Fourth & Count \\
        \midrule
        restaurant	&	train	&	-	&	-	&	87	\\
        attraction	&	train	&	-	&	-	&	80	\\
        hotel	&	-	&	-	&	-	&	71	\\
        train	&	attraction	&	-	&	-	&	71	\\
        train	&	hotel	&	-	&	-	&	70	\\
        restaurant	&	-	&	-	&	-	&	64	\\
        train	&	restaurant	&	-	&	-	&	62	\\
        hotel	&	train	&	-	&	-	&	57	\\
        taxi	&	-	&	-	&	-	&	51	\\
        attraction	&	restaurant	&	-	&	-	&	38	\\
        restaurant	&	attraction	&	taxi	&	-	&	\textbf{35}	\\
        restaurant	&	attraction	&	-	&	-	&	31	\\
        train	&	-	&	-	&	-	&	31	\\
        hotel	&	attraction	&	-	&	-	&	27	\\
        restaurant	&	hotel	&	-	&	-	&	27	\\
        restaurant	&	hotel	&	taxi	&	-	&	\textbf{26}	\\
        attraction	&	hotel	&	taxi	&	-	&	\textbf{24}	\\
        attraction	&	restaurant	&	taxi	&	-	&	\textbf{23}	\\
        hotel	&	restaurant	&	-	&	-	&	22	\\
        attraction	&	hotel	&	-	&	-	&	20	\\
        hotel	&	attraction	&	taxi	&	-	&	\textbf{16}	\\
        hotel	&	restaurant	&	taxi	&	-	&	\textbf{13}	\\
        attraction	&	-	&	-	&	-	&	12	\\
        attraction	&	restaurant	&	train	&	-	&	3	\\
        restaurant	&	hotel	&	train	&	-	&	3	\\
        hotel	&	train	&	restaurant	&	-	&	3	\\
        restaurant	&	train	&	hotel	&	-	&	3	\\
        restaurant	&	taxi	&	hotel	&	-	&	3	\\
        attraction	&	train	&	restaurant	&	-	&	2	\\
        train	&	attraction	&	restaurant	&	-	&	2	\\
        attraction	&	restaurant	&	hotel	&	-	&	2	\\
        hotel	&	train	&	attraction	&	-	&	2	\\
        attraction	&	taxi	&	hotel	&	-	&	1	\\
        hotel	&	taxi	&	-	&	-	&	1	\\
        train	&	hotel	&	restaurant	&	-	&	1	\\
        restaurant	&	taxi	&	-	&	-	&	1	\\
        restaurant	&	train	&	taxi	&	-	&	\textbf{1}	\\
        hotel	&	restaurant	&	train	&	-	&	1	\\
        hotel	&	taxi	&	train	&	-	&	1	\\
        taxi	&	attraction	&	-	&	-	&	1	\\
        restaurant	&	train	&	attraction	&	-	&	1	\\
        attraction	&	train	&	hotel	&	-	&	1	\\
        attraction	&	train	&	taxi	&	-	&	\textbf{1}	\\
        restaurant	&	attraction	&	train	&	-	&	1	\\
        hotel	&	taxi	&	attraction	&	-	&	1	\\
        train	&	hotel	&	attraction	&	-	&	1	\\
        restaurant	&	taxi	&	attraction	&	-	&	1	\\
        hotel	&	attraction	&	restaurant	&	taxi	&	\textbf{1}	\\
        attraction	&	hotel	&	train	&	-	&	1	\\
        taxi	&	restaurant	&	train	&	-	&	1	\\
        \bottomrule
    \end{tabularx}
    \end{threeparttable}
\end{table}
\newpage

\textcolor{white}{.}
\newpage
\section{Inference Time Complexity (ITC)}
\label{appendix:itc}

\input{tabs/itc}


\noindent
Inference Time Complexity (ITC) proposed by \citet{ren2019comer} defines the efficiency of a DST model using $J$, the number of slots, and $M$, the number of values of a slot.
Going a step further from their work, we report ITC of the models in the best case and the worst case for relatively more precise comparison.

Table \ref{table:itc} shows ITC of several models in their best and worst cases.
Since our model generates values for only the slots on which \operation{update} operation has to be performed, the best case complexity of our model is $\Omega(1)$, when there is no slot whose operation is \operation{update}.

\newpage
\section{Sample Outputs}
\label{appendix:examples}

\begin{figure*}[!h] 
    \centering
    \includegraphics[width=\textwidth]{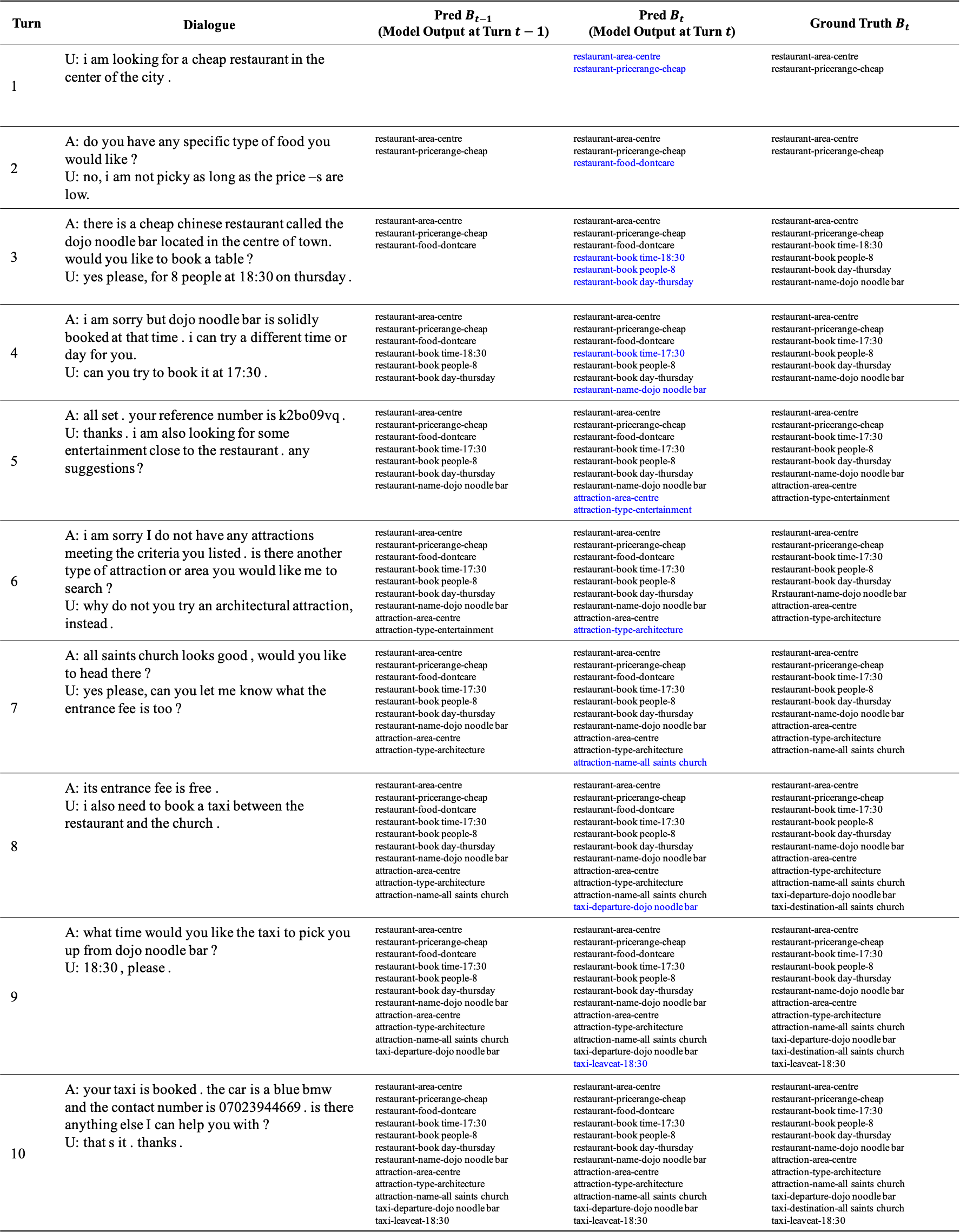}
    \caption{
        The output of \ours{} in a dialogue (\texttt{dialogue\_idx} MUL2499) in the test set of MultiWOZ 2.1. Parts changed from the previous dialogue state are shown in blue. To save space, we omit the slots with value \texttt{NULL} from the figure.
    }
    \label{fig:ex1}
\end{figure*}

\begin{figure*}[!h] 
    \centering
    \includegraphics[width=\textwidth]{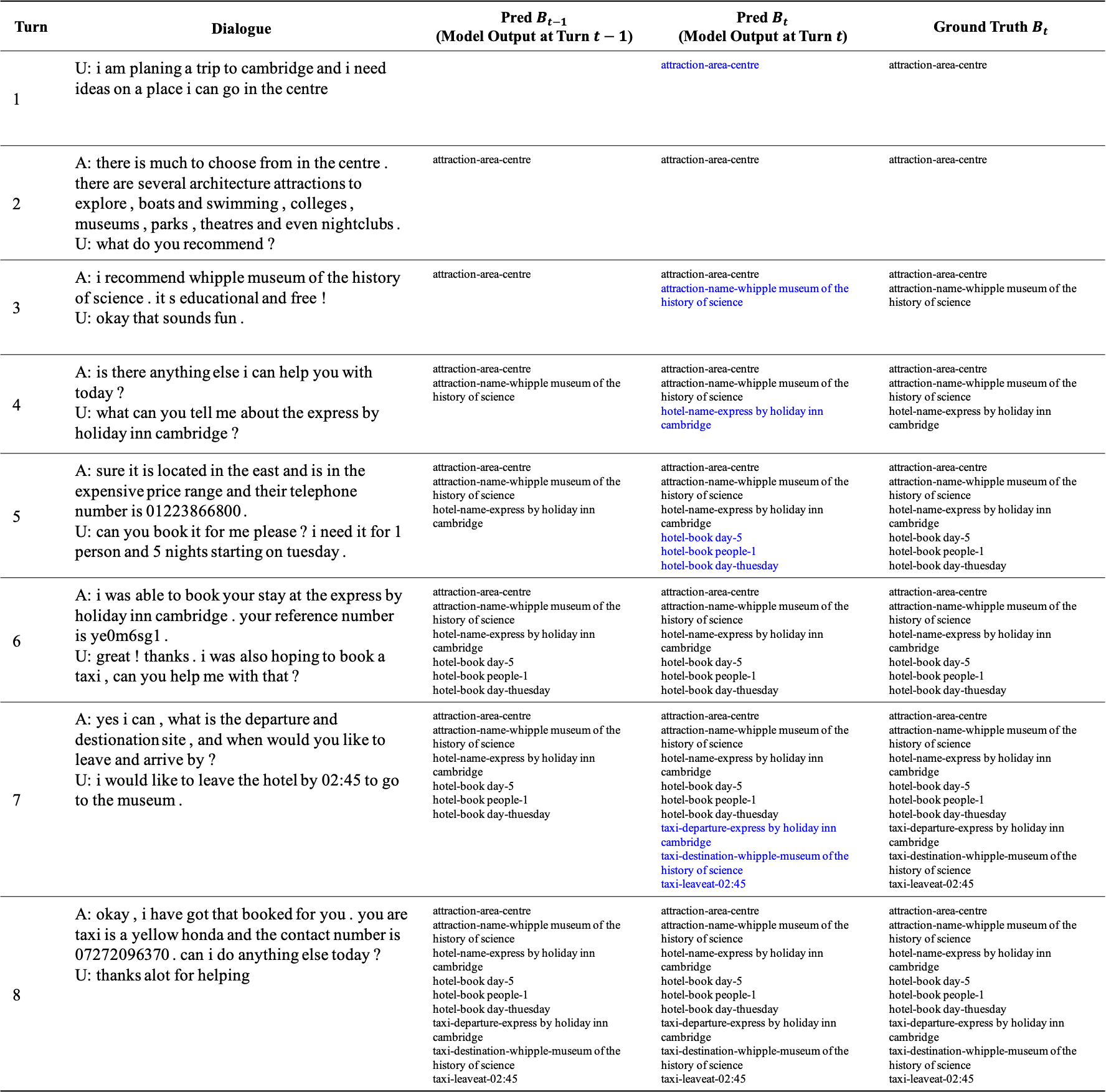}
    \caption{
        The output of \ours{} in a dialogue (\texttt{dialogue\_idx} PMUL3748) in the test set of MultiWOZ 2.1. Parts changed from the previous dialogue state are shown in blue. To save space, we omit the slots with value \texttt{NULL} from the figure.
    }
    \label{fig:ex2}
\end{figure*}

\end{appendices}

%% file: tabs/itc.tex
\begin{table}[h!]
    \centering
    \caption{Inference Time Complexity (ITC) of each model. We report the ITC in both the best case and the worst case for more precise comparison. $J$ indicates the number of slots, and $M$ indicates the number of values of a slot.}
    \label{table:itc}
    \footnotesize
    \begin{threeparttable}
    \begin{tabular*}{\columnwidth}{l@{\extracolsep{\fill}}ll} 
        \toprule
            & \multicolumn{2}{c}{Inference Time Complexity} \\
        \cmidrule{2-3}
        Model           & Best           & Worst          \\ 
        \midrule
        SUMBT             & $\Omega(JM)$ & $O(JM)$ \\ 
        DS-DST            & $\Omega(J)$  & $O(JM)$ \\ 
        DST-picklist      & $\Omega(JM)$ & $O(JM)$ \\ 
        DST Reader        & $\Omega(1)$  & $O(J)$  \\ 
        TRADE             & $\Omega(J)$  & $O(J)$  \\ 
        COMER             & $\Omega(1)$  & $O(J)$  \\ 
        NADST             & $\Omega(1)$  & $O(1)$  \\ 
        ML-BST            & $\Omega(J)$  & $O(J)$  \\ 
        \ours{}(ours)     & $\Omega(1)$  & $O(J)$  \\ 
        \bottomrule
    \end{tabular*}
    \end{threeparttable}
\end{table}